\documentclass[sigconf]{acmart}
\usepackage{multicol, multirow}
\usepackage[shortlabels]{enumitem}
\usepackage{pervasives}
\usepackage{framed}
\usepackage{balance}
\usepackage{booktabs}
\usepackage{arydshln}
\usepackage{xspace}
\usepackage{soul}
\usepackage{xcolor}

\newcommand{\ours}{\textsc{Seq2Seq-Ptr}\xspace}

\AtBeginDocument{%
  \providecommand\BibTeX{{%
    \normalfont B\kern-0.5em{\scshape i\kern-0.25em b}\kern-0.8em\TeX}}}

\setcopyright{acmcopyright}
\copyrightyear{2020}
\acmYear{2020}
\setcopyright{iw3c2w3}
\acmConference[WWW '20]{Proceedings of The Web Conference 2020}{April 20--24, 2020}{Taipei, Taiwan}
\acmBooktitle{Proceedings of The Web Conference 2020 (WWW '20), April 20--24, 2020, Taipei, Taiwan}
\acmPrice{}
\acmDOI{10.1145/3366423.3380064}
\acmISBN{978-1-4503-7023-3/20/04}

\begin{document}

\title[Don't Parse, Generate! A Sequence to Sequence Architecture for Task-Oriented Semantic Parsing]{Don't Parse, Generate! A Sequence to Sequence \\ Architecture for Task-Oriented Semantic Parsing}

\author{Subendhu Rongali}
\authornote{Work done when the author was at Amazon Alexa AI for a summer internship.}
\affiliation{%
  \institution{University of Massachusetts Amherst}
  \city{Amherst}
  \state{MA}
  \state{USA}
  \postcode{01002}
}
\email{srongali@cs.umass.edu}

\author{Luca Soldaini}
\affiliation{%
  \institution{Amazon Alexa Search}
  \city{Manhattan Beach}
  \state{CA}
  \country{USA}}
\email{lssoldai@amazon.com}

\author{Emilio Monti}
\affiliation{%
  \institution{Amazon Alexa}
  \city{Cambridge}
  \country{UK}}
\email{monti@amazon.co.uk}

\author{Wael Hamza}
\affiliation{%
  \institution{Amazon Alexa AI}
  \city{New York}
  \state{NY}
  \country{USA}}
\email{waelhamz@amazon.com}

\newcommand\todo[1]{\texttt{\small \textcolor{red}{#1}}}
\newcommand\subbu[1]{\texttt{\small \textcolor{orange}{#1}}}

\begin{abstract}
Virtual assistants such as Amazon Alexa, Apple Siri, and Google Assistant often rely on a semantic parsing component to understand which action(s) to execute for an utterance spoken by its users. 
Traditionally, rule-based or statistical slot-filling systems have been used to parse ``simple'' queries; that is, queries that contain a single action and can be decomposed into a set of non-overlapping entities. 
More recently, shift-reduce parsers have been proposed to process more complex utterances. 
These methods, while powerful, impose specific limitations on the type of queries that can be parsed; namely, they require a query to be representable as a parse tree. 
  
In this work, we propose a unified architecture based on Sequence to Sequence models and Pointer Generator Network to handle both simple and complex queries. 
Unlike other works, our approach does not impose any restriction on the semantic parse schema. 
Furthermore, experiments show that it achieves state of the art performance on three publicly available datasets (ATIS, SNIPS, Facebook TOP), relatively improving between 3.3\% and 7.7\% in exact match accuracy over previous systems.
Finally, we show the effectiveness of our approach on two internal datasets. 
\end{abstract}



\keywords{Natural Language Understanding, Semantic Parsing, Voice Assistants, Sequence to Sequence models}


\maketitle

\section{Introduction}
Adoption of intelligent voice assistants such as Amazon Alexa, Apple Siri, and Google Assistant has increased dramatically among consumers in the past few years:
as of early 2019, it is estimated that 21\% of U.S. adults own a smart speaker, a 78\% year-over-year growth \cite{npr2019smart}.
These systems are built to process user dialog and perform tasks such as media playback and online shopping. 

A major part of any voice assistant is a semantic parsing component designed to understand the action requested by its users:
given the transcription of an utterance, a voice assistant must identify the action requested by a user (play music, turn on lights, etc.), as well as parse any entities that further refine the action to perform (which song to play? which lights to turn on?)
Despite huge advances in the field of Natural Language Processing (NLP), this task still remains challenging due to the sheer number of possible combinations a user can use to express a command. 

\begin{figure}
    \centering
    \includegraphics[width=1\columnwidth]{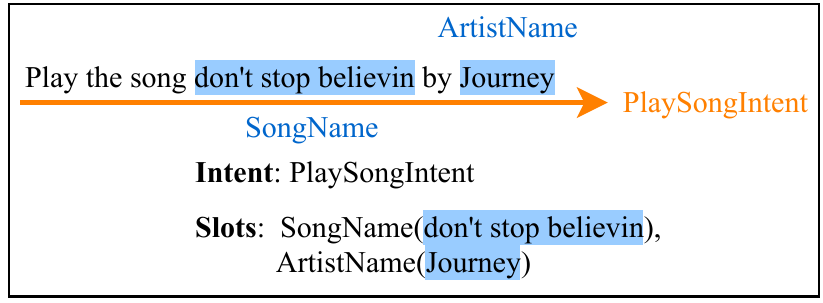}
    \caption{Semantic parsing of a ``simple'' query. 
    Simple queries define single action (intent) and can be decomposed into a set of non-overlapping entities (slots).}
    \label{fig:sem_par}
\vspace{-1em}
\end{figure}

Traditional approaches for task-oriented semantic dialog parsing frame the problem as a slot filling task. 
For example, given the query \emph{Play the song don't stop believin by Journey}, a traditional slot filling system parses it in two independent steps: 
(\textit{i}) It first classifies the \emph{intent} of the user utterance as \texttt{PlaySongIntent}, and then (\textit{ii}) identifies relevant \emph{named entities} and tags those slots, such as \emph{don't stop believin} as a \texttt{SongName} and \emph{Journey} as an \texttt{ArtistName}. 
Traditional semantic parsing can therefore be reduced to a text classification and a sequence tagging problem, which is a standard architecture for many proposed approaches in literature \cite{liu2016attention,mesnil2013investigation,lafferty2001conditional}. This is shown in Figure \ref{fig:sem_par}.

With increasing expectations of users from virtual assistants, there is a need for the systems to handle more complex queries -- ones that are composed of multiple intents and nested slots or contain conditional logic. 
For example, the query \emph{Are there any movie in the park events nearby?} involves first finding the location of \emph{parks} that are \emph{nearby} and then finding relevant \emph{movie} events in them. 
This is not straightforward in traditional slot filling systems. \citet{gupta2018semantic} and \citet{einolghozati2019improving} proposed multiple approaches for this using a Shift-reduce parser based on Recurrent Neural Network Grammars \cite{dyer2016recurrent} that performs the tagging. 

In this paper, we propose a unified approach to tackle semantic parsing for natural language understanding based on Transformer Sequence to Sequence models \cite{vaswani2017attention} and a Pointer Generator Network \cite{vinyals2015pointer,see2017get}. 
Furthermore, we demonstrate how our approach can leverage pre-trained resources, such as neural language models, to achieve state of the art performance on several datasets.
In particular, we obtain relative improvements between 3.3\% and 7.7\% over the best single systems on three public datasets (SNIPS \cite{coucke2018snips}, ATIS \cite{price1990evaluation} and TOP \cite{gupta2018semantic}, the last consisting of complex queries); on two internal datasets, we show relative improvements of up to 4.9\%.

Furthermore, our architecture can be easily used to parse queries that do not conform to the grammar of either the slot filling or RNNG systems. Some examples include semantic entities that correspond to overlapping spans in the query, and entities comprising of non-consecutive spans. We do not report any results on these kinds of datasets but we explain how to formulate the problems using our architecture.

In summary, our contributions are as follows.
\begin{itemize}
    \item We propose a new architecture based on Sequence to Sequence models and a Pointer Generator Network to solve the task of semantic parsing for understanding user queries.
    \item We describe how to formulate different kinds of queries in our architecture. Our formulation is unified across queries with different kinds of tagging.
    \item We achieve state-of-the-art results on three public datasets and two internal datasets.
\end{itemize}

\begin{figure}
    \centering
    \includegraphics[width=1.0\columnwidth]{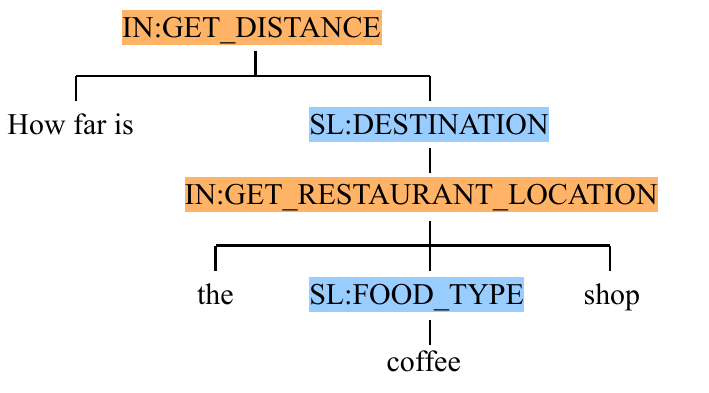}
    \caption{Semantic parse for a ``complex'' query in the Facebook TOP dataset. This complex query is represented as a tree containing two nested intents and slots.}
    \label{fig:top}
\vspace{-1em}
\end{figure}

\section{Methodology}

\begin{figure*}
    \centering
    \includegraphics[width=1\textwidth]{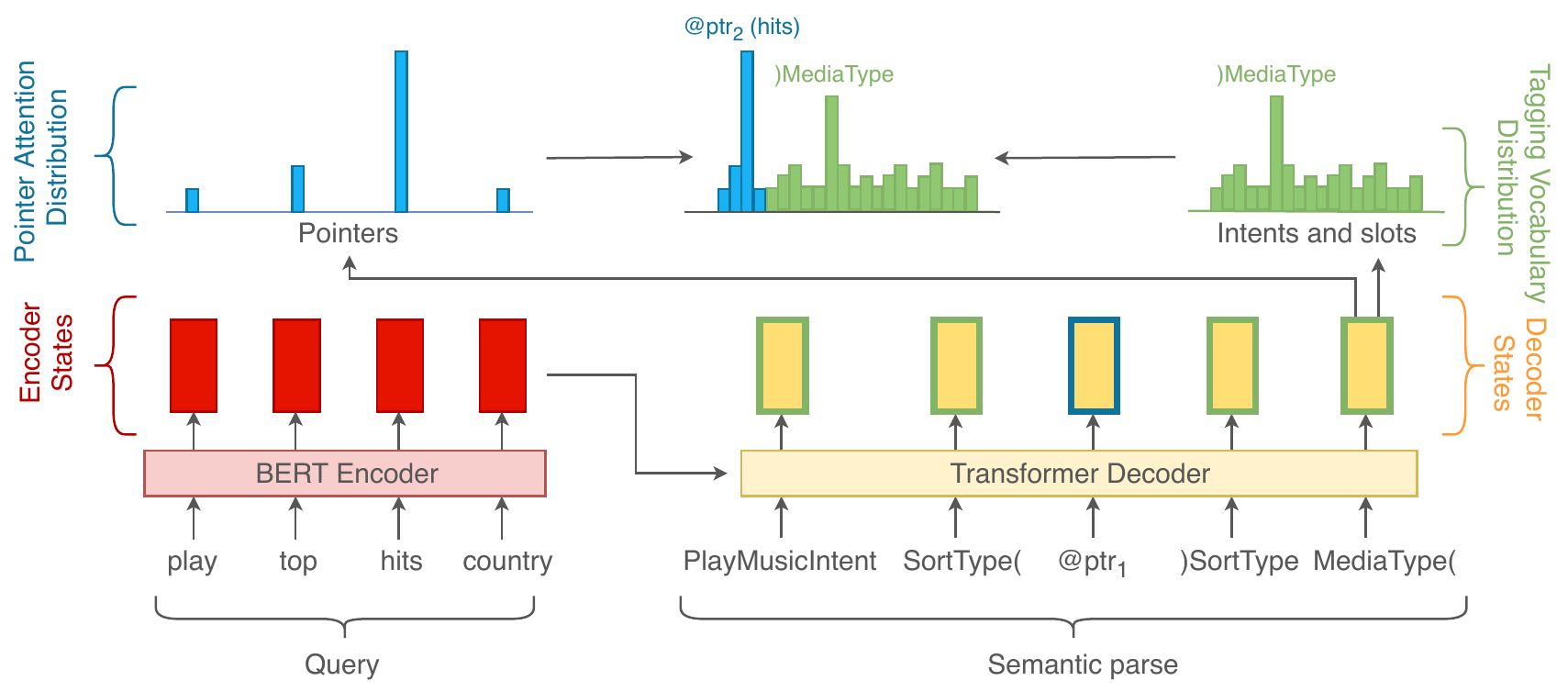}
    \caption{Our architecture - Sequence to Sequence model with Pointer Generator Network (\ours). The model is currently decoding the symbol after \texttt{MediaType(} by looking at the scores over the tagging vocabulary and the attentions over the source pointers. It generates \texttt{$@ptr_2$} since it has the highest overall score.}
    \label{fig:arch}
\end{figure*}

We propose a unified architecture to solve the task of semantic parsing for both simple and complex queries. This architecture can also be adapted to handle queries containing slots with overlapping spans. It consists of a Sequence to Sequence model and a Pointer Generator Network. We choose a pretrained BERT \cite{devlin2018bert} model as our encoder. Our decoder is modeled after the transformer decoder described in \citet{vaswani2017attention} and is augmented with a Pointer Generator Network \cite{vinyals2015pointer,jia2016data} which allows us to learn to generate pointers to the source sequence in our target sequence. 
Figure \ref{fig:arch} shows this architecture parsing an example query. We train the model using a cross-entropy loss function with label smoothing.

In this section, we first describe how we formulate queries and their semantic parses as sequences with pointers for our architecture. We then describe our encoder and decoder components.

\subsection{Query Formulation}
A Sequence to Sequence architecture is trained on samples with a source sequence and a target sequence. When some words in the target sequence are contained in the source sequence, they can be replaced with a separate \emph{pointer} token that points to that word in the source to be able to apply the Pointer Generator Network.

Take the example query from Figure \ref{fig:sem_par}. In our architecture, we use the query as our source sequence. The target sequence is constructed by combining the intent with all the slots, in order, with each slot also containing its source words. The source and target sequences now look as follows.

\begin{lstlisting}
Source: play the song don't stop believin by journey
Target: PlaySongIntent SongName( $@ptr_3$ $@ptr_4$ $@ptr_5$ 
        )SongName ArtistName( $@ptr_7$ )ArtistName
\end{lstlisting}

Here, each token \texttt{$@ptr_i$} is a pointer to the $i^{th}$ word in the source sequence. So \texttt{$@ptr_3$}, \texttt{$@ptr_4$} and \texttt{$@ptr_5$} point to the song words \emph{don't stop believin}, and \texttt{$@ptr_7$} points to the artist word \emph{journey}. The slots have open and close tags since they are enclosing a consecutive span of source tokens. The intent is just represented as a single tag at the beginning of the target sequence. We can do this for simple queries since they consist of just one intent. The target vocabulary hence consists of all the available intents, two times the number of different slots, and the pointers.

Complex queries with multiple intents and nested slots can also be transformed easily into this formulation. Figure \ref{fig:top} shows an example from the Facebook TOP dataset along with its parse tree. This query \emph{How far is the coffee shop} can be converted into our formulation as follows.

\begin{lstlisting}
Source: How far is the coffee shop
Target: [IN:GET_DISTANCE $@ptr_0$ $@ptr_1$ $@ptr_2$ [SL:DESTINATION 
    [IN:GET_RESTAURANT_LOCATION $@ptr_3$ [SL:TYPE_FOOD 
    $@ptr_4$ SL:TYPE_FOOD] $@ptr_5$ IN:GET_RESTAURANT_LOCATION]
    SL:DESTINATION] IN:GET_DISTANCE]
\end{lstlisting}


We made a minor modification to the reference parses from the TOP dataset for our formulation. We replaced the end-brackets with custom end-brackets corresponding to the intent or slot they close. We found that this formulation helped our models perform better.

Finally, we show how we can express queries from datasets that don't conform to either the slot-filling or Shift-reduce systems. Take the following example from the healthcare domain, where the task is to extract a patient diagnosis and related information from a clinician's notes.

\begin{lstlisting}
Source: The pt. was diagnosed with GI upper bleed today.
Annotations: Bleeding_Event (GI bleed),
             Anatomical_Site (upper)
\end{lstlisting}

A traditional slot filling system wouldn't know which non consecutive slots to combine, while a shift-reduce parser cannot split the middle word into a separate tag. In our architecture, we simply formulate the target sequence as follows.

\begin{lstlisting}
Target: Bleeding_Event( $@ptr_5$ $@ptr_7$ )Bleeding_Event
        Anatomical_Site( $@ptr_6$ )Anatomical_Site
\end{lstlisting}

\subsection{BERT Encoder}
Language model pretraining has been shown to improve the downstream performance on many NLP tasks \cite{peters2018deep,radford2019language,devlin2018bert}. The idea is to train a language model on a large amount of text using a next word prediction objective to learn good representations for each of the words. These representations can then be fine-tuned on a given NLP task to improve the performance of an existing model. Pretrained models improve the performance of task models since they already contain a lot of useful semantic information learned through the pretraining phase. This has even more significance when the task-specific dataset is fairly small. Some examples of pretrained models in literature include word embeddings such as Word2Vec \cite{mikolov2013distributed} and Glove \cite{pennington2014glove}, and contextualized representations such as ELMo \cite{peters2018deep}, OpenAI-GPT \cite{radford2019language}, and BERT \cite{devlin2018bert}.

We choose BERT to encode the source sequence in our architecture. BERT (Bidirectional Encoder Representations from Transformers) is a language representation model architecture based on Transformers \cite{vaswani2017attention}. 
The original publicly available model was pretrained on a millions of lines of text from BooksCorpus and English Wikipedia.
Unlike other language models (ELMo, OpenAI-GPT), which are trained to predict the next token given the previous sequence of words, BERT uses a composite objective that combines masked word prediction and next sentence prediction. 

BERT's architecture is based on a multi-layer bidirectional Transformer, originally implemented in \citet{vaswani2017attention}. The detailed implementation of this architecture can be found in \citet{devlin2018bert}. For our experiments, we use three different variants of BERT. 

For the three public datasets, we used the checkpoint released by \citet{devlin2018bert}. Experiments on the two internal datasets were carried out using a model we pretrained over a large sample of queries from the live traffic of Amazon Alexa. 
We also experimented with a publicly-available variant of BERT called RoBERTa \cite{liu2019roberta}. RoBERTa (A Robustly Optimized BERT Pretraining Approach) uses the same architecture as BERT but changes the pretraining process. The next sentence prediction objective is removed and a dynamic masking scheme is used instead of a static one like the original BERT implementation. RoBERTa was also trained with longer sequences, higher batch-sizes, and for a longer time, and was reported to match or exceed the performance of BERT in several NLP benchmarks. The detailed implementation can be found in \citet{liu2019roberta}.
Finally as an ablation study, we experimented with an encoder with no pretrained weights. 

\subsection{Decoder with Pointer Generator Network}

We use the transformer decoder proposed in \citet{vaswani2017attention} in our architecture. The self-attention mechanism in the decoder learns to attend to target words before the current step, as well as all the source words in the encoder. 

We set up the decoder with different numbers of units, layers, and attention-heads for different tasks based on the size and complexity of the queries. These details are provided in the experiments section.

In a traditional Sequence to Sequence model, the target words are generated from the decoder hidden states through a feed-forward layer that obtains unnormalized scores over a target vocabulary distribution. In our architecture, we use a Pointer Generator Network to generate two different kinds of target words: words from the target vocabulary consisting of parse symbols (the intent and slot delimiters), and words that are simply pointers to the source sequence. Our Pointer Generator Network is based on the models in \citet{vinyals2015pointer} and \citet{see2017get}, and is closest in implementation to \citet{jia2016data}.


We now describe our decoding process. For each input source sequence $[x_1 \dots x_n]$, we use the BERT encoder to encode it into a sequence of encoder hidden states $[e_1 \dots e_n]$. Having generated the first $t-1$ output tokens, the transformer decoder generates the token at step $t$ as follows. 

First, the decoder produces the decoder hidden state at time $t$, $d_t$ by building multi-layer multi-head self-attention on the encoded output as well as the embeddings of the previously generated output sequence as described in \citet{vaswani2017attention}. We feed $d_t$ through a dense layer to produce scores $[s_1, \dots s_{|V|}]$ for each word in the vocabulary $V$. $V$ contains all symbols necessary for the parse (intents, slots) but not regular words appearing on the source side.

We also use $d_t$ as a query and compute unnormalized attention scores $[a_1 \dots a_n]$ with the encoded sequence. Concatenating the unnormalized attention scores (size $n$) and the output of the dense layer (size $|V|$), we obtain an unnormalized distribution over $|V| + n$ tokens, the first $|V|$ of which are the output parsing vocabulary and the last $n$ of which are the \texttt{$@ptr_i$} ($0<i<n$) words pointing to the source tokens. We then feed this through a softmax layer to obtain the final probability distribution. This probability is used in the loss function during training and will be used to choose the next token to generate during inference. Since the transformer decoder uses embeddings of previously generated tokens, we use a set of special embeddings to represent \texttt{$@ptr_i$} tokens.

In the example in Figure \ref{fig:arch}, we are trying to predict a target word after the token \texttt{'MediaType('} at step 5. As shown in the figure, we compute the scores $[a_1 \dots a_4]$ (blue, left) over each of the source tokens, and the scores $[s_1 \dots s_{|V|}]$ (green, right) over the parsing vocabulary. We expect the model to produce the highest score for $a_3$, which corresponds to \texttt{$@ptr_2$}, representing the word \emph{hits}.

\section{Datasets}


We test our approach on five different datasets (three publicly available, two interal), which we describe in this section.  



\subsection{Facebook TOP}

The Task Oriented Parsing (TOP) \cite{gupta2018semantic} dataset contains complex hierarchical and nested queries that make the task of semantic parsing more challenging. It contains around 45k annotations with 25 intents and 36 slots, randomly split into 31k training, 5k validation and 9k test utterances. The dataset mainly consists of user queries about navigation and various public events. An example from this dataset can be seen in Figure \ref{fig:top}. 
The \texttt{IN:} prefix stands for intent while \texttt{SL:} is for slot. We can see how there are multiple intents and nested slots in the semantic interpretation. This makes the query much harder to interpret and parse using a simple slot tagging model that tags each word with a single slot.

\subsection{SNIPS}
The SNIPS dataset \cite{coucke2018snips} is a public dataset that is used for training and testing semantic parsing models for voice assistants. It consists of utterances that belong to seven different intents: SearchCreativeWork, GetWeather, BookRestaurant, PlayMusic, AddToPlaylist, RateBook, and SearchScreeningEvent. Each intent contains around 2000 examples to train and 100 to test.

This dataset contains only simple queries with single intents and flat slots. An example is \emph{Will there be fog in Tahquamenon Falls State Park}, where the intent is \texttt{GetWeather} and the slots are \texttt{condition\_description} for \emph{fog} and \texttt{geographic\_poi} for \emph{Tahquamenon Falls State Park}.


The dataset was originally used to evaluate models in the Snips Voice Platform. It has since been a widely used dataset to benchmark the performance of various task-oriented parsing models.

\subsection{ATIS}


The Airline Travel Information System (ATIS) \cite{price1990evaluation} corpus is a widely used dataset in spoken language understanding. It was built by collecting and transcribing audio recordings of people making flight reservations in the early 90s. It consists of simple queries.
 
There are seventeen different goals or intents such as \texttt{Flight} or \texttt{Aircraft capacity}. This distribution is however skewed, with the \texttt{Flight} intent covering about 70\% of the total queries. An example from this dataset consists of the query \emph{How much is the cheapest flight from Boston to New York tomorrow morning?} The intent is \texttt{Airfare}, while the slots tag important information like the departure and arrival cities, and the departure times.
 
The ATIS corpus has supported research in the field of spoken language understanding for more than twenty years. Some researchers have performed extensive error analysis on the state of the art discriminative models for this dataset and reported that despite really low error rates, there exist many unseen categories and sequences in the dataset that can benefit from incorporating linguistically motivated features \cite{tur2010left}. This supports the continued utility of ATIS as a research corpus.

\subsection{Internal Datasets}
\label{sub:int_data}

Our internal datasets consist of millions of user utterances that are used to train and test Amazon Alexa. 
For our experiments, we sampled two datasets of utterances, one from the music domain and the other from video domain. 
Utterances in these domains naturally included a large amount of entities (e.g. artists and albums names, movie and video titles), and thus represent a good benchmark for the ability of any neural model to generalize over a diverse set of queries. The example in Figure \ref{fig:arch} is from the music domain.





The sampled music domain dataset contains 6.2M training and 200k test utterances, with 23 intents and 100 slots. 
The video domain dataset contains 1M training and just 5k test utterances; 
parses in this dataset are comprised of 24 distinct intents and 59 slots. 

\section{Baseline Models}

We benchmark our performance on the internal datasets by comparing it to a well tuned RNN based model. 
The model learns to perform joint intent and slot tagging using a bidirectional LSTM and a Conditional Random Field (CRF) \cite{huang2015bidirectional}.   
We further enhanced this baseline by replacing its embedding and encoder layers with a language model pretrained on a subset of Alexa's live traffic.
These components were fine-tuned on the two datasets described in Section~\ref{sub:int_data}.

For the ATIS and SNIPS datasets, we use the top four performing methods reported by \citet{zhang2019joint} as baselines. All these models perform joint intent and slot tagging. There are two variants that use RNNs: a simple RNN based model, and an RNN model augmented with attention. There is also a model that works completely with just attention, the slot gated full attention model. The final baseline, CapsuleNLU, uses Capsule Networks \cite{sabour2017dynamic}.

For the TOP dataset, we pick a model based on Recurrent Neural Network Grammars (RNNG) \cite{dyer2016recurrent}, the Shift Reduce Parser. We provide a brief overview of this model as described in \citet{gupta2018semantic} - the parse tree is constructed using a sequence of transitions, or \emph{actions}. The transitions are defined as a set of SHIFT, REDUCE, and the generation of intent and slot labels. SHIFT action consumes an input token (that is, adds the token as a child of the right most \emph{open} sub-tree node) and REDUCE closes a sub-tree. The third set of actions is generating non-terminals: the slot and intent labels. The model learns to perform one of these actions at each step in time. 

We report scores of three experimental setups with the shift reduce parser from \citet{einolghozati2019improving}: a simple shift reduce parser, a shift reduce parser augmented with ELMo embeddings, and an ensemble of these models augmented with ELMo and an SVM language model reranker. 

\begin{table}[t]
\begin{tabular}{ccc}
\toprule
\multirow{2}{*}{\bf{Method}} & \multicolumn{2}{c}{\bf{Accuracy}} \\
& \bf{exact match} & \bf{intent} \\ 
\midrule
Shift Reduce (SR) Parser \cite{einolghozati2019improving} & 80.86 & -- \\
SR with ELMo embeddings \cite{einolghozati2019improving} & 83.93 & -- \\
{\small SR ensemble + ELMo + SVMRank} \cite{einolghozati2019improving} & \bf{87.25} & -- \\
\hdashline
\ours (no pretraining) & 79.25 & 97.43 \\
\ours (BERT encoder) & 83.13 & 97.91 \\
{\ours (RoBERTa encoder)} & 86.67 & \bf{98.13} \\
\bottomrule
\end{tabular}
\caption{
    Results on \textbf{TOP} \citep{gupta2018semantic}.
    Our system (\ours) outperforms the best single method (SR + ELMo) by 3.3\%. It is close to the best ensemble approach (SR + ELMo + SVRank). 
}
\label{tab:top}
\vspace{-2em}
\end{table}

\begin{table}[t]
\begin{tabular}{ccc}
\toprule
\multirow{2}{*}{\bf{Method}} & \multicolumn{2}{c}{\bf{Accuracy}} \\
& \bf{exact match} & \bf{intent} \\ 
\midrule
Joint BiRNN \cite{hakkani2016multi} & 73.20 & 96.90 \\
Attention BiRNN \citep{liu2016attention} & 74.10 & 96.70 \\
Slot Gated Full Attention \citep{goo2018slot} & 75.50 & 97.00 \\
\textsc{CapsuleNLU} \citep{zhang2019joint} & 80.90 & 97.30 \\
\hdashline
\ours (no pretraining) & 85.43 & 97.00 \\
\ours (BERT encoder) & 86.29 & \bf{98.29} \\
{\ours (RoBERTa encoder)} & \bf{87.14} & {98.00} \\
\bottomrule
\end{tabular}
\caption{
    Results on \textbf{SNIPS} \citep{coucke2018snips}.
    Our system (\ours) outperforms the previous state of the art by 7.7\%. 
}
\label{tab:snips}
\vspace{-2em}
\end{table}

\begin{table}
\begin{tabular}{ccc}
\toprule
\multirow{2}{*}{\bf{Method}} & \multicolumn{2}{c}{\bf{Accuracy}} \\
& \bf{exact match} & \bf{intent} \\ 
\midrule
Joint BiRNN \cite{hakkani2016multi} & 80.70 & 92.60 \\
Attention BiRNN \citep{liu2016attention} & 78.90 & 91.10 \\
Slot Gated Full Attention \citep{goo2018slot} & 82.20 & 93.60 \\
\textsc{CapsuleNLU} \citep{zhang2019joint} & 83.40 & 95.00 \\
\hdashline
\ours (no pretraining) & 81.08 & 95.18 \\
\ours (BERT encoder) & 86.37 & 97.42  \\
{\ours (RoBERTa encoder)} & \bf{87.12} & \bf{97.42} \\
\bottomrule
\end{tabular}
\caption{
    Results on \textbf{ATIS} \cite{price1990evaluation}.
    Our system (\ours) outperforms the previous best method by 4.5\%.
}
\label{tab:atis}
\vspace{-2em}
\end{table}

\section{Experimental Setup}

All our models were trained on a machine with 8 NVIDIA Tesla V100 GPUs, each with 16GB of memory. 
When using pretrained encoders, we leveraged gradual unfreezing to effectively tune the language model layers on our datasets. 
We used the "\textsc{Base}" variant of BERT and RoBERTa encoders, which uses 768-dimensional embeddings, 12 layers, 12 heads, and 3072 hidden units. 
When training from scratch, we used a smaller encoder consisting of 512-dimensional embeddings, 6 layers, 8 heads, and 1024 hidden units.

Depending on the dataset, we used either a 128 units, 4 layers, 3 heads, and 512 hidden units decoder (Facebook TOP, ATIS, SNIPS) or a larger 512 units, 6 layers, 8 heads, and 1024 hidden units decoder (internal Music and Video datasets). 
We used bi-linear product attention to score the source words in the Pointer Network. 

While training, the cross entropy loss function was modified with label smoothing with $\epsilon = 0.1$. We used the Adam \cite{kingma2014adam} optimizer with noam learning rate schedule \cite{vaswani2017attention}, each adjusted differently for different datasets. 
At inference time, we used beam search decoding with a beam size of 4. 

\section{Results and Discussion}

We use exact match (EM) accuracy as the main metric to measure the performance of our models across all datasets. 
Under this metric, the entire semantic parse for a query has to match the reference parse to be counted as correct. 
Because EM is generally more challenging than slot-level precision and recall or semantic error rate \cite{thomson2012nbest}, it is better suited to compare high performing systems like the ones studied in this work.
For completeness, we also report the intent classification accuracy for our models.


The results from our experiments are documented in Tables \ref{tab:top}-\ref{tab:intvid}. Our models match or beat the baselines across all datasets on both exact match and intent classification accuracies. 
We see significant improvements on both simple and complex datasets. 

\subsection{Complex Queries}

We achieve an improvement of 2.7 (+3.3\%) EM accuracy points on the TOP dataset over the state-of-the-art single model on this dataset (Table~\ref{tab:top}).
Our \ours model with RoBERTa encoder is only surpassed by the ensemble model reported in \citet{einolghozati2019improving} (+0.6\% EM accuracy points.) 

In addition, we find that even without specifying any hard requirements for the grammar of the parse trees in the complex queries, 98\% of the generated parses are well formatted. For the simple query datasets, it was greater than 99\% but difference is expected since the grammar is easier to learn there. 

During error analysis, we found an interesting example in the TOP dataset where we believe our model generates a valid, more meaningful parse than the reference annotation. For the query \emph{What time do I need to leave to get to Helen by 8pm}, our model parses \emph{Helen} as  \texttt{[SL:DESTINATION [IN:GET\_} \texttt{LOCATION\_HOME [SL:CONTACT Helen ] ] ]}, while it is annotated as \texttt{[SL:DESTINATION Helen ]}.
Our parse resolves the query as finding the estimated departure time to get to a location that is the home location of a contact named Helen, while the reference annotation suggests that the correct interpretation is to 
find the estimated departure time to get to a destination named \emph{Helen}.
We believe our parse is more likely to be correct given that Helen is most likely the name of a person. 

\subsection{Simple Queries}
We report results of our sequence to sequence model (\ours) on four datasets (SNIPS, ATIS, internal music, internal video) that contain simple queries in Tables \ref{tab:snips}, \ref{tab:atis}, \ref{tab:intmus}, and \ref{tab:intvid}.

On the SNIPS and ATIS datasets, we note that the best version of our method (\ours with RoBERTa encoder) achieves a significant improvement in EM accuracy over existing baselines (+7.7\% and +4.5\% respectively.)
Using a BERT encoder causes a slight decrease in performance, but still achieves a meaningful improvement over the previous state of the art \cite{zhang2019joint};
this is consistent with what has been observed on other NLP tasks \cite{liu2019roberta}.
If no pretraining is used, performance is further reduced but it is notable that this variant still beats all the baselines on the SNIPS dataset.

For our internal Alexa datasets, we note that the proposed \ours method obtains comparable results to a BiLSTM-CRF tagger on the music domain, and slightly better EM accuracy (+1.9\%) on the video domain.
We believe our model wasn't able to outperform the baseline on the music domain because the entities in this domain are very diverse, especially song or album names. Sequence tagging methods therefore benefit from having to solve a simpler task of having to tag each word in the sequence, as opposed to our unconstrained model.
We would however like to note that our from-scratch variants beat the from-scratch baselines on both domains.
Also curiously, the performance of \ours with a BERT encoder fell behind that of a sequence to sequence model trained from scratch.
Since the scratch model uses a smaller transformer encoder (6 layers with 8 heads per layer instead of 12/12), we believe it was able to converge more effectively than the BERT encoder.



\begin{table}[t]
\begin{tabular}{ccc}
\toprule
\multirow{2}{*}{\bf{Method}} & \multicolumn{2}{c}{\bf{Accuracy}} \\
& \bf{exact match} & \bf{intent} \\ 
\midrule
BiLSTM-CRF (no pretraining) & \multicolumn{2}{c}{\textit{baseline}} \\
BiLSTM-CRF (pretrained LM) & \bf{+3.0\%} & \bf{+0.1\%} \\
\hdashline
\ours (no pretraining) & -0.3\% & -0.7\%  \\
\ours (BERT encoder) & -2.2\% & -0.8\%  \\
{\ours (RoBERTa encoder)} & {-3.5\%} & {-0.7\%}\\
\bottomrule
\end{tabular}
\caption{Results on an internal dataset (music). The best configuration of our method (\ours) is comparable to a BiLSTM-CRF pretrained on a large conversational dataset.}
\label{tab:intmus}
\vspace{-2em}
\end{table}

\begin{table}[t]
\begin{tabular}{ccc}
\toprule
\multirow{2}{*}{\bf{Method}} & \multicolumn{2}{c}{\bf{Accuracy}} \\
& \bf{exact match} & \bf{intent} \\ 
\midrule
BiLSTM-CRF (no pretraining) & \multicolumn{2}{c}{\textit{baseline}} \\
BiLSTM-CRF (pretrained LM) & +3.0\% & \bf{+0.1\%} \\
\hdashline
\ours (no pretraining) & +2.9\% & -0.1\%  \\
\ours (BERT encoder) & +0.1\% & -0.2\%  \\
{\ours (RoBERTa encoder)} & \bf{+4.9\%} & {-0.2\%} \\
\bottomrule
\end{tabular}
\caption{Results on an internal dataset (video). The best configuration of our method (\ours) outperforms a pretrained BiLSTM-CRF network by 1.9\%.}
\label{tab:intvid}
\vspace{-2em}
\end{table}

\section{Related Work}


The task of semantic parsing for intent and slot detection is well established in literature. Traditionally, this was done with slot filling systems that classify the query and then label each word in the query. There were a few approaches that followed this system, using Recurrent Neural Networks \cite{liu2016attention,mesnil2013investigation}. Researchers have also experimented with Convolutional Neural Networks and showed good results \cite{kim2014convolutional} and more recently, Capsule Networks \cite{sabour2017dynamic,zhang2019joint}. 

Prior to the advent of deep learning models, the task of sequence labeling was tackled with the use of Conditional Random Fields (CRF) \cite{lafferty2001conditional,jiao2006semi,peng2004chinese}. CRFs learn pairwise potentials on labeling subsequent words which allow models to find more probable label sequences for a given query.

Most of this work is valid for semantic parsing for simple queries which boils down to a sequence labeling task. To handle more complex cases with hierarchical slots such as the example in Figure \ref{fig:top}, researchers have experimented with Sequence to Sequence models and models based on Recurrent Neural Network Grammars (RNNG) \cite{dyer2016recurrent}. RNNGs were shown to perform better on complex queries than RNN or Transformer-based Sequence to Sequence models \cite{gupta2018semantic}. Researchers have also explored models involving logical forms and discourse for language representation \cite{liang2016learning,zettlemoyer2012learning,van2018exploring}. 


The Pointer Generator Network in our architecture was introduced in \citet{vinyals2015pointer}. It was used in NLP applications where some words from the source sequence reappeared in the target sequence such as text summarization and style transfer \cite{see2017get,paulus2017deep,prabhumoye2018style}. They were also used to copy out of vocabulary words from the source to target in machine translation \cite{klein2017opennmt}. Our implementation of the Pointer Network is closest to the architecture in \citet{jia2016data}. By using pointers to represent the source tokens and imposing no particular logical form over our target sequence, we can handle any kind of queries for parsing. This makes our architecture as expressive as logical forms, while also being able to learn as easily as simple slot tagging systems.

\section{Conclusion}
We propose a unified architecture for the task of semantic parsing for different kinds of queries. We show that our architecture matches or outperforms existing approaches across multiple datasets : internal music and video datasets, SNIPS, ATIS, and Facebook TOP. We significantly outperform the current state of the art models on the public datasets TOP (3.3\%), SNIPS (7.7\%), and ATIS (4.5\%). 

We describe how to apply this architecture to both simple queries and complex queries with hierarchical and nested slots. We also describe how to formulate any set of queries with non-conforming grammars to work with our architecture, making this model applicable to many different types of semantic parsing. We leave the non-conforming grammar task to future work. 
\bibliographystyle{ACM-Reference-Format}
\bibliography{main}










\end{document}